# Position Estimation of Camera Based on Unsupervised Learning

YanTong Wu

School of Digital Media & Design Arts,
Beijing University of Posts and
Telecommunications, Beijing, 100876

wuyantong@bupt.edu.cn

Yang Liu

School of Digital Media & Design Arts,
Beijing University of Posts and
Telecommunications, Beijing, 100876

yang.liu@bupt.edu.cn

## ABSTRACT

It is an exciting task to recover the scene's 3D structure and camera pose from the video sequence. Most of the current solutions divide it into two parts, monocular depth recovery and camera pose estimation. The monocular depth recovery is often studied as an independent part, and a better depth estimation is used to solve the pose. While camera pose is still estimated by traditional SLAM (Simultaneous Localization And Mapping) methods in most cases. The use of unsupervised method for monocular depth recovery and pose estimation has benefited from the study of [1] and achieved good results. In this paper, we improve the method of [1]. Our emphasis is laid on the improvement of the idea and related theory, introducing a more reasonable inter frame constraints and finally synthesize the camera trajectory with inter frame pose estimation in the unified world coordinate system. And our results get better performance.

## 1. INTRODUCTION

Recovery of 3D structure of scene and camera pose estimation from video attracts plenty of interest for its widely usage in Robotics, Augmented Reality, etc. It is often divided into two parts: depth recovery and camera pose estimation. Deep learning is dominating the Computer Vision, most state-of-the-art depth recovery algorithms benefit from CNN (Convolutional Neural Network), while researchers prefer to traditional methods to estimate the pose of camera.

For the monocular depth recovery, like [5, 6, 7] generally use a supervised CNN to learn the pixel depth. Pixel depth can be transformed into spatial 3D coordinates, thus providing initial values for pose estimation. However, the final coordinates of the scene's points need to be unified with the camera pose. Only relying on better pixel depth estimation does not guarantee this. At present, most studies apply deep learning to SLAM to view the task independently and hope to provide better initial estimation for pose solving by training a better depth recovery network. So for the camera pose estimation, most algorithm still adopt the traditional method through feature extraction, matching and tracking like [10, 11, 14] or through direct method like [2, 3].

Zhou [1] proposed to recover the 3D structure of scene and estimate the camera pose by deep learning in an unsupervised manner which shows a great potential in this direction. The network in [1] is divided into two parts, DispNet and Pose network. The DispNet is mainly used for monocular depth recovery, and Pose network outputs the adjacent frames position and posture. Using the output of depth recovery network as the input of the pose network, the two networks are trained together through the constraints of light intensity between adjacent frames and getting a good result. However [1] only constraints that the light intensity of all the corresponding points between adjacent frames remains unchanged. We suggest to introduce more reasonable constraints in order to improve the performance. What's more, this method only considers the pose between adjacent frames, which has not been normalized to the unified coordinate system.

Our contributions is two-fold:

1. Introducing stronger inter frame constrains to make the network structure more reasonable. We add the number of projected frames in calculating the light intensity error, and introduce gradient –light-intensity error (details in chapter 3.3).

2. Deriving the path synthesis method from the inter frame output to the unified coordinate system. [1] only output the inter frame pose, and not splicing them into a camera trajectory.

## 2. RELATED WORK

Monocular SLAM is one of the most attractive schemes in various SLAM solution and it is also one of the most important research topics in the field of computer vision. Its goal is to reconstruct the 3D structure and camera position of the scene from the video sequence. Its two important problems are the Monocular depth recovery and the inter frame pose estimation

### 2.1 Depth estimation

For monocular depth recovery, most of them are trained using supervised models, such as [5, 7], and the network structure generally has a contractive process (corresponding to the process of convolution, for example, the step length of the convolution kernel is set to 2, then the output's size after the convolution is 1/4 of the original), and then there will be a expansion part corresponding to the deconvolution process(the network structure is sampled in the [5, 7, 12, 13, 16]), and the final loss function will be depth error between the output and ground truth. In general, in order to keep the continuity of the depth map, smooth loss (such as [5, 12, 13, 16]) is added.

### 2.2 Pose estimation

For inter frame pose estimation, most solution still use traditional methods. In traditional SLAM methods, the calculation of pose is often used as the back-end of the system. Pose is optimized by constructing the re-projection error between Inter frames using the method of graph optimization [15]. The method generally involves features extraction and matching, and the accuracy of pose estimation often depends on the result of matching points to a large extent. And there are other methods not involving the features extraction and matching like LSD-SLAM [2, 3]. It is solved directly by the constraints of the light intensity of corresponding points between adjacent frames. In detail, it is divided into the dense method and sparsity method based on the points' source involved in the calculation. No matter which solution, the process of pose calculation and the building of the graph are carried out at the same time. Generally the method will choose key frames, and assume points are subject to the Gauss distribution (such as [2, 3], the inverse depth is initialized in the case of small parallax, the proof process comes from [18]) or the mixed Gauss distribution (such as [14]). The inter frame pose is computed in turn to calculate the coordinates of the spatial points and project them to the current frame for depth fusion. It can be said that for the traditional SLAM, the pose estimation and mapping process are constrained each other. The results of pose estimation will be used to update the points in the map, and the points in the map will be used to calculate the pose of the next frame, and the two processes are carried out simultaneously.

## 2.3 Unsupervised learning for SLAM

The application of deep learning to the research of SLAM often separates the monocular depth recovery process from the pose estimation, and often concentrates on training a better depth recovery network to estimate inter frames pose using traditional methods such as [9]. There is little special research on the learning of pose networks. Even if there is exists, it is just an auxiliary study to the main task such as STN (spatial transformer network) in [8]. The network is embedded in the traditional classification network. In order to improve the accuracy of the classification, a STN is added to make affine transformation to the image originally crooked.

In this case, SLAM methods based on depth learning are rarely used in unsupervised way. Purely using unsupervised way to do SLAM profit from the research of [1]. It connects the monocular depth recovery network and the pose estimation network, and finally constructs a minimum inter frame error to train the entire depth recovery network and pose estimation network. In theory, there are constraints in the whole scene. As long as we can find a stronger description of the unified constraints of the scene, we can get better inter frame pose and point depth. The models obtained by deep learning often tend to have better generalization ability and can avoid the weakening of performance caused by artificial feature selection. Furthermore, due to the inherent convenience of unsupervised learning, it will be one of the key contents for the future SLAM research.

## 3. OUR APPROACH

Our network structure is still CNN. It can be divided into two parts: monocular depth recovery network and pose estimation network. The monocular depth network not only outputs the depth corresponding to the input image, but also outputs the place where the image gradient is more obvious. In particular, the input image is a continuous three frame image sequence, and output is the depth of each frame, and Laplace operator acting on the input frames is used to get the image gradient, see the details in chapter 3.1. For the pose network, input is the depth map of the three frame sequence from the output of the monocular depth recovery network, and outputs the pose between adjacent frames. Details are shown in chapter 3.2. Finally, the inter frame light intensity constraint is constructed as the power of updating parameters for the whole network. Details are shown in chapter 3.3.

## 3.1 Monocular depth recovery network
### 3.1.1 Network structure

Our network structure still adopts the form of CNN (using the structure like [5, 7, 12, 13, 16]). The convolution process is equivalent to the extraction process of the depth feature. The larger convolution kernel and less convolution kernel number are used in the lower layer, and the smaller convolution kernel and more convolution kernel number are used in the higher level, which is beneficial to the network using lower cost on the learning of shallow features, and using higher cost on the more abstract features.

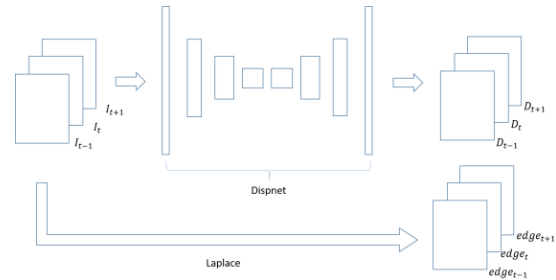

**Figure 1. Monocular depth recovery network. The input of the network is an image sequence with three frames connected, and the output is the corresponding depth and edge of each frame. Edge extraction use Laplace operator. The monocular recovery network is Dispnet [1]. The number of rectangles in the graph does not represent the number of convolution layers, just to show the structure of the network.**

In practice, the size of convolution kernel is 7*7, 5*5, 3*3, 3*3, 3*3 as the depth of the layer increases, and respectively the number of convolution kernel is 32, 64, 128, 256, 512. The deconvolution part correspond to the convolution layer before, the kernel size and number are the same. The structure diagram is as shown in Figure 1.

### 3.1.2 High-gradient points

The points with high gradient are often the points on the edge of the image. So we can use traditional method to get the edge of an input image. Here we choose Laplace operator, as shown in Figure 2. The Laplace operator is applied to the input image, and a reasonable threshold is obtained to get the obvious gradient of the input image. The place where the image gradient is obvious is used to modify the premise that the light intensity of the corresponding points in adjacent frames are constant. We think that the corresponding points on the edge of the image will have an important effect on the calculation of rotation and translation.

The significance of getting the points on the edge is to modify their weights. This idea is mainly derived from LSD-SLAM [2]. In LSD-SLAM, the author constructs the illumination intensity error of the corresponding points between adjacent frames. An image coordinates point $p_1$ in the previous frame $I_1$, through the reverse projection, rotation and translation, finally project to the point $p_2$ in the next frame $I_2$. By minimizing the intensity error between the point $p_1$ in $I_1$ and the point $p_2$ in $I_2$ to optimize the rotation and translation between the two frames. In its final formula derivation, it can be seen that the derivative of light intensity error is made up of two parts. First one is the derivative of

$I_2$ to the coordinate of point $p_2$, and second one is the derivative of $p_2$ to the rotation and translation matrix. The first part is actually the gradient of the image $I_2$. So the rotation and translation between the two frames is closely related to the gradient of $I_2$. It is precisely because of this that LSD-SLAM is divided into dense and sparse two methods according to the source of points involved in the calculation, and the sparse method only uses the points with high gradient.

So our idea is to increase the weight of the points with high gradient and make these important points align as much as possible. And the results of LSD-SLAM prove that these points have an important impact on the final pose.

| 0 | 1 | 0 |
|---|---|---|
| 1 | -4 | 1 |
| 0 | 1 | 0 |

| 1 | 1 | 1 |
|---|---|---|
| 1 | -8 | 1 |
| 1 | 1 | 1 |

**Figure 2. Laplace operator. It is used to process the input image and get the points with high gradient.**

### 3.2 Pose estimation network

This part mainly outputs the pose of the frame *t-1* and the frame *t+1* relative to the frame *t* [1]. Since the pose transformation between the two adjacent frames is determined by rotation and translation, and the rotation is an orthogonal matrix, which is essentially determined by 3 parameters (the angle of rotation along the X axis in the current camera coordinate system, the angle of the rotation along the Y axis and the Z axis, here we respectively name them α, β, γ), and the translation is determined by 3 parameters too (here we respectively name them $T_x$, $T_y$, $T_z$). Finally the rotation and translation transformation between every two frames is determined by 6 parameters, so our network output is a 2*6 matrix, each row represents the rotation and translation of the frame *t-1* and the frame *t+1* relative to the frame *t* respectively. In addition, this network is implemented with a pure convolution neural network with the 8 convolution layers. The kernel size are 7*7, 5*5, 3*3, 3*3, 3*3, 3*3, 3*3, 1*12.

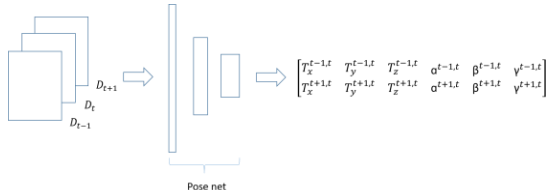

**Figure 3 Position estimation network. The input of the network is the depth image sequence with three continuous frames. And $T_x^{t-1,t}$ represents the translation of the camera position of the frame *t* relative to the frame *t-1* in the direction of the X axis, and $α^{t-1,t}$ represents the angle of rotation along X axis of the frame *t* relative to the frame *t-1*, and so on.**

The translation and three Eulerian angles produced by the pose estimation network can generate the rotation and translation matrix of the frame *t* relative to the frame *t-1* and the frame *t* relative to the frame *t+1* (the dimension is 4*4 in the form of homogeneous coordinate).

### 3.3 Loss function
#### 3.3.1 Light intensity error

Because our goal is to projection two frame in the *t-1*, *t*, *t+1* frame to another frame and minimize the light intensity of its corresponding points. The projection frame can be *t-1* or *t* or t+1, so the minimum light intensity error is composed by three parts. Details are as follows:

(1) The target is the frame *t*, then the loss is:

$$L_t = \sum_{t\in[1,2,\ldots,N-1]}(\sum_{p_1}|I_{t-1}(p_1) - \hat{I}_t(p_2)| + \sum_{p_1}|I_{t+1}(p_1) - \hat{I}_t(p_2)|) \quad (1)$$

The above formula mainly describes the sum of light intensity error of projection from the frame *t-1* to the frame *t* and the projection of the frame *t+1* to the frame *t*. Taking the projection of the frame *t-1* to the frame *t* as an example, $p_1$ is the homogeneous coordinate of the frame *t-1*, and $p_2$ is the homogeneous coordinate of its corresponding point in the frame *t*, then there are:

$$p_2 = KT_{t-1,t}D_{t-1}(p_1)K^{-1}p_1 \quad (2)$$

In the formula (2), $K$ is the camera's inner parameter matrix, which is considered to be a known quantity. And $T_{t-1,t}$ is the rotation and translation of the frame *t* relative to the frame *t-1*. It is the rotation and translation matrix generated by the output matrix's first row of the pose network (generated by the Euler rotation formula and the homogeneous coordinate), and $D_{t-1}(p_1)$ is the depth value of the point $p_1$ in frame *t-1*.

Similarly, if we project the frame *t+1* to the frame *t*, then (2) will be the following form:

$$p_2 = KT_{t+1,t}D_{t+1}(p_1)K^{-1}p_1 \quad (3)$$

And $T_{t+1,t}$ is the rotation and translation of the frame *t* relative to the frame *t+1*. That is the rotation and translation matrix generated by the output matrix's second row of the pose network. $\hat{I}_t(p_2)$ is the result after bilinear interpolation (adopted in [1, 8]), because the coordinates of $p_2$ may not be integers after transformation. The latter description is similar to above.

(2) The target is the frame *t-1*, then the loss is:

$$L_{t-1} = \sum_{t\in[1,2,\ldots,N-1]}(\sum_{p_1}|I_t(p_1) - \hat{I}_{t-1}(p_2)| + \sum_{p_1}|I_{t+1}(p_1) - \hat{I}_{t-1}(p_2)|) \quad (4)$$

Similarly, for the homogeneous coordinate point $p_1$ in the frame *t*, the corresponding point $p_2$ in the frame *t-1* has the following form:

$$p_2 = KT_{t,t-1}D_t(p_1)K^{-1}p_1$$
$$T_{t,t-1} = T_{t-1,t}^{-1} \quad (5)$$

For the homogeneous coordinate point $p_1$ in the frame *t+1*, the corresponding point $p_2$ in the frame *t-1* has the following form:

$$p_2 = KT_{t+1,t-1}D_{t+1}(p_1)K^{-1}p_1$$
$$T_{t+1,t-1} = T_{t-1,t}^{-1}T_{t+1,t} \quad (6)$$

(3) The target is the frame *t+1*, then the loss is:

---
[1] Our input is the continuous three frames image, here for convenience we name the middle frame *t* and the previous one frame *t-1* and the latter one frame *t+1*.

$$L_{t+1} = \sum_{t\in[1,2,...,N-1]}(\sum_{p_1}|I_{t-1}(p_1) - \hat{I}_{t+1}(p_2)| + \sum_{p_1}|I_t(p_1) - \hat{I}_{t+1}(p_2)|) \quad (7)$$

For the homogeneous coordinate point $p_1$ in the frame *t-1*, the corresponding point $p_2$ in the frame *t+1* has the following form:

$$p_2 = KT_{t-1,t+1}D_{t-1}(p_1)K^{-1}p_1$$
$$T_{t-1,t+1} = T_{t+1,t}^{-1}T_{t-1,t} \quad (8)$$

Similarly, for the homogeneous coordinate point $p_1$ in the frame *t*, the corresponding point $p_2$ in the frame *t+1* has the following form:

$$p_2 = KT_{t,t+1}D_t(p_1)K^{-1}p_1$$
$$T_{t,t+1} = T_{t+1,t}^{-1} \quad (9)$$

In addition, in order to improve continuity of the depth image we add smooth loss (just like [1, 6]), so the total intensity error is

$$L_{intensity} = L_t + L_{t-1} + L_{t+1} + \lambda_s L_{smooth} \quad (10)$$

### 3.3.2 Gradient –light-intensity error

In addition to the common light intensity errors mentioned above, we propose the concept of gradient–light-intensity error. As described in the sparse method of traditional SLAM, the solution of pose is largely related to the point with high gradient. For the iterative method in solving convex optimization problems, the derivative of certain point needs to be well passed down to make a greater contribution to convergence, so that the flat region of the image where the gradient is not obviously will cause gradient loss. Sparsity method has done it, so we have a reason to believe that points with high gradient should have more important effects, so they should be treated differently from the ordinary points and increase their weight. Under such idea the gradient–light-intensity error is as follows:

$$L_{edge} = \sum_{t\in[1,2,...,N-1]}(\sum_{p_1\in edge(I_{t-1})}|I_{t-1}(p_1) - \hat{I}_t(p_2)| + \sum_{p_1\in edge(I_{t+1})}|I_{t+1}(p_1) - \hat{I}_t(p_2)|) \quad (11)$$

### 3.3.3 Final loss

According to the above, our loss function consists of two parts, one is the light intensity error between all the corresponding points, and the other is the light intensity error of the points with high gradient. In addition, we think that the point with high gradient is more important to iterative solution of the pose, so their corresponding coefficient should be not less than 1. Total loss function is as follows:

$$L_{final} = L_{intensity} + \lambda_e L_{edge} \quad (\lambda_e \geq 1) \quad (12)$$

## 4. EXPERIMENT

We use the dataset on KITTI to train our network, using the framework of open source tensorflow (version 1.2), taking 09 and 10 sequences as tests, and comparing the results with paper [1]. This chapter mainly focuses on model parameters selection and experimental results.

### 4.1 Model parameters

For the specific training process, first of all, our dataset uses 9.26, 9.28, 10.3 (part) three day's data on KITTI about 15000 image sequences for training. Before training, we need to splice the image into every three consecutive sequence. As the parameters in [1], the input image is scaled to 128*416 for training, the batch size is 4, the soomth weight is set to 0.5, the iterative method is still Adam optimizer with beta=0.9, the learning rate is 0.0002, the activation function is Relu. The whole training process does not use the pre-training model.

Our experimental results show that it is better to use the 4 neighborhood Laplace operator to extract obvious gradient points. For hyper parameter $\lambda_e$, we do some experiment to choose the better one. Because our training error is about 0.7 in the case of $\lambda_e = 0$, and in order to increase the weight of the gradient part, the $\lambda_e$ should be greater than 1, so we tested the $\lambda_e$ from 100 to 1 every 10, and get the results as Figure 4.

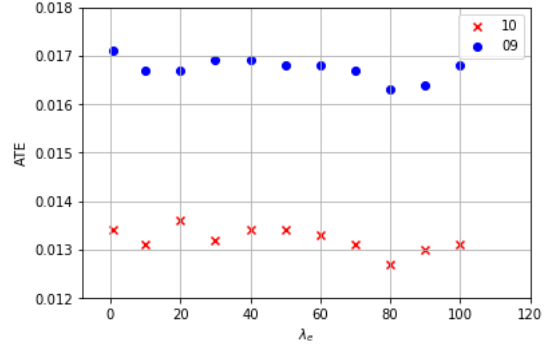

**Figure 4. ATE of sequence 09 and 10 in different $\lambda_e$.**

The results shows that the ATE (Absolute Trajectory Error) will be better when $\lambda_e$ takes 10 or 80. However, from the results of the final camera trajectory (see chapter 4.2) we get better results with $\lambda_e=20$ or $\lambda_e=80$. So the best ATE is not necessarily the best model to trajectory synthesis. How to select better $\lambda_e$ is still our future work.

### 4.2 Experimental results

For the pose estimation test, we use 09 and 10 sequences on KITTI. Since our network output is the pose transformation between three adjacent frames, in order to avoid introducing the extra error, we cut the ground truth of the 09 and 10 sequences' pose into every three continuous frames, and calculate the ATE error in every three continuous frames (According to the standard of [1]). The final results are as Table 1:

**Table 1. Absolute Trajectory Error (ATE) on the KITTI odometry split averaged over all 3-frame snippets (lower is better). We use the same input setting, and we can get our method is better than [1] and ORB-SLAM (short).**

| Method | Seq.09 | Seq.10 |
| --- | --- | --- |
| ORB-SLAM(full) | 0.014±0.008 | 0.012±0.011 |
| ORB-SLAM(short) | 0.064±0.141 | 0.064±0.130 |
| Tinghui Zhou &al.[1] | 0.021±0.017 | 0.020±0.015 |
| Ours | 0.0163±0.0059 | 0.0127±0.0081 |

From the experimental results, we can see that by increasing the number of projection frames and increasing the weight of the gradient part, both the mean and variance of ATE are significantly reduced.

Further, in order to see the visual effect of the trajectory, we stitching the trajectories between the three adjacent frames of the output (concrete steps in the appendix), unifying it to the same coordinate system. Finally we splicing our final camera trajectory and the ground truth into the same world coordinate system through the method [17]. The final results are shown as Figure 5:

From the results of the final track stitching, our results are superior to the results in [1] taking the sequence 09 for example. Our overall trajectory shape is more consistent with the ground truth with $\lambda_e$=20 or $\lambda_e$=80. When $\lambda_e$ takes 20, it still shows a good result even though it's not good at numerical value on ATE. Although we didn't consider loop detection like traditional SLAM method, our model still performs well in the last ring, which shows the strong vitality of unsupervised learning in camera pose estimation. It is worth mentioning that our path synthesis method is sequential (details in appendix) and that may introduce extra error, and it is also a follow-up study to splice the trajectories of adjacent frames better.

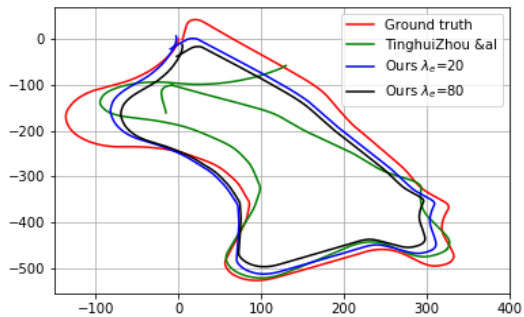

**Figure 5. Camera trajectory of sequence 09 on KITTI.**

## 5. DISCUSSION

Our overall idea is to minimize the illumination intensity error using a stronger inter frame constraints. On the other hand, we should treat different pairs of corresponding points differently and allocate different weights to different points. Since the work of LSD-SLAM [2, 3], we believe a more significant point with higher gradient should get more important weight, that is to say, in addition to all the points shall match well, the more important points (where the gradient is high) should do better. In this way, the rotation and translation obtained through the iterations should be better.

Further, our ideas can be extended to other directions, such as semantic SLAM. We still divide the point into two categories, common points and corresponding points between adjacent frames of the same objects. We shouldn't only minimize the light intensity error of the whole corresponding points, but also consider the points' quality supported by the prior knowledge. One possible way is increasing the weights of the corresponding points on the same object. Sometimes the object takes up a region, there could be flat surfaces preventing the backpropagation of gradient, and we can consider the weights are decreasing under a certain distribution centered a key point. The final purpose is to make the more important points play more important role. Multiple view geometry [4] shows that in extreme cases, rotation and translation calculated by 4 points is the most accurate, if the corresponding points between the two frames match without any error. Adding other points will reduce the accuracy. Finally how to allocate weights reasonably is the direction of our future study.

# APPENDIX

This part mainly introduces the process of sequentially splicing three consecutive pose sequences into a completed trajectory. Suppose $X_k$ (k=0, 1, 2 …, n) is camera pose completed by the final splicing, then in frame $k$ we output the pose $T'_{k,k+1}$ for frame $k+1$ relative to frame $k$, and the pose $T'_{k,k+1}$ for frame $k+2$ relative to frame $k$. Below is discussing how to transform the relative inter pose to the same coordinate system with $X_0$ as the coordinate origin.

First, it is necessary to calculate the inter frame translation according to the inter frame pose matrix. The argument is as follows: Here, we take the translation of the frame $k+1$ relative to the frame $k$ for example. Suppose $P_1$ is a space point with three dimension in the camera coordinates corresponding to the frame $k$. $P_2$ is its corresponding point in the camera coordinate system of the frame $k+1$. And the rotation of the camera $k+1$ relative to the camera $k$ is recorded as $R_{k,k+1}$, and the translation is recorded as $t_{k,k+1}$, then we have

$$R_{k,k+1}(P_1 - t_{k,k+1}) = P_2$$
$$\Rightarrow \begin{bmatrix} R_{k,k+1} & -R_{k,k+1}t_{k,k+1} \\ 0^T & 1 \end{bmatrix} \begin{bmatrix} P_1 \\ 1 \end{bmatrix} = \begin{bmatrix} P_2 \\ 1 \end{bmatrix} \quad (1)$$
$$\Leftrightarrow T'_{k,k+1} \begin{bmatrix} P_1 \\ 1 \end{bmatrix} = \begin{bmatrix} P_2 \\ 1 \end{bmatrix}$$

We use $t'_{k,k+1}$ to stand for the translation separated from $T'_{k,k+1}$ (the vector composed of the first 3 rows in the last column), then get:

$$t_{k,k+1} = -R_{k,k+1}^{-1} t'_{k,k+1} \quad (2)$$

In the same way we can get the relative translation $t_{k,k+2}$, however we should keep in mind that the relative translation is in the local coordinate system. What we should get is the translation in the global coordinate system, so our goal is:

$$t_{0,k} = -R_{0,k}^{-1} t'_{0,k} \quad (3)$$

From (3), we can know if we want to get the best global translation $t_{0,k}$, we should optimize $R_{0,k}^{-1}$ and $t'_{0,k}$. For convenience, we use unit quaternion $q_{k,k+1}$ to express $R_{k,k+1}$, so our goal to optimize the rotation matrix will be optimizing $q_{0,k}$. For the camera $k$, we can optimize it through measurement respectively in the frame $k-1$ and $k-2$. So we can get the following deduction:

$$\begin{cases} q_{0,k}^{(1)} = q_{k-2,k} q_{0,k-2} \\ q_{0,k}^{(2)} = q_{k-1,k} q_{0,k-1} \\ q_{0,k} = Slerp(q_{0,k}^{(1)}, q_{0,k}^{(2)}) \end{cases} \quad (4)$$

In (4), $q_{0,k}^{(1)}$ stands for optimizing result of first measurement in the frame $k-2$, $q_{0,k}^{(2)}$ stands for optimizing result of second measurement in the frame $k-1$, $Slerp$ is a function for quaternion interpolation (in the experiment, we use the middle interpolation).

Similarly, we can get the method of optimizing to $t'_{0,k}$, for convenience, $t'_{0,k}$ in the following derivation formula we can think it a vector with 4 dimension ( the last dimension is 0 for quaternion multiplication).

$$\begin{cases} t'^{(1)}_{0,k} = q_{k-2,k} t'_{0,k-2} q_{k-2,k}^{-1} + t'_{k-2,k} \\ t'^{(2)}_{0,k} = q_{k-1,k} t'_{0,k-1} q_{k-1,k}^{-1} + t'_{k-1,k} \\ t'_{0,k} = (t'^{(1)}_{0,k} + t'^{(2)}_{0,k})/2 \end{cases} \quad (5)$$

Through the iterative calculations of (4) and (5), we can get the final $R_{0,k}$ and $t'_{0,k}$. And through (3) we can get the final global translation $t_{0,k}$.